\documentclass[journal]{IEEEtran}
\usepackage{cite}

\ifCLASSINFOpdf
\else
\fi
\usepackage{graphicx}
\usepackage{subfigure}
\usepackage{multirow}
\usepackage{booktabs}
\usepackage[numbers]{natbib}
\usepackage{amsmath}
\usepackage[switch]{lineno}
\usepackage{algorithm}
\usepackage{algorithmic}
\usepackage{float}
\usepackage{graphicx}
\usepackage{subfigure}
\usepackage{amsmath}
\usepackage{color}
\usepackage{array}

\usepackage{threeparttable}
\begin{document}

%
\title{Image Captioning via a Hierarchical Attention Mechanism and Policy Gradient Optimization}
%
%
%

\author{Shiyang~Yan,
        Yuan Xie,
        Fangyu Wu,
        Jeremy~S. Smith,~\IEEEmembership{Member,~IEEE,}
        Wenjin~Lu 
        and~Bailing~Zhang
\thanks{Shiyang Yan, Yuan Xie and Bailing Zhang are with the Institute of Advanced Artificial Intelligence in Nanjing, Nanjing, China}
\thanks{Jeremy S. Smith is with the University of Liverpool.}
\thanks{Fangyu Wu, Wenjin Lu are with the Department
of Computer Science and Software Engineering, Xi'an Jiaotong-Liverpool University, Suzhou, China.}
\thanks{Manuscript received; revised}}

%
%

\markboth{Journal of \LaTeX\ Class Files,~Vol.~14, No.~8, August~2015}%
{Shell \MakeLowercase{\textit{et al.}}: Bare Demo of IEEEtran.cls for IEEE Journals}
%



\maketitle

\begin{abstract}
Automatically generating the descriptions of an image, i.e., image captioning, is an important and fundamental topic in artificial intelligence, which bridges the gap between computer vision and natural language processing. Based on the successful deep learning models, especially the CNN model and Long Short-Term Memories (LSTMs) with attention mechanism, we propose a hierarchical attention model by utilizing both of the global CNN features and the local object features for more effective feature representation and reasoning in image captioning. The generative adversarial network (GAN), together with a reinforcement learning (RL) algorithm, is applied to solve the exposure bias problem in RNN-based supervised training for language problems. In addition, through the automatic measurement of the consistency between the generated caption and the image content by the discriminator in the GAN framework and RL optimization, we make the finally generated sentences more accurate and natural. Comprehensive experiments show the improved performance of the hierarchical attention mechanism and the effectiveness of our RL-based optimization method. Our model achieves state-of-the-art results on several important metrics in the MSCOCO dataset, using only greedy inference.
\end{abstract}


%
\IEEEpeerreviewmaketitle

\section{Introduction}

Naturalistic description of an image is one of the primary goals of computer vision, which has recently received much attention in the field of artificial intelligence recently. It is a high-level task and much more complicated than some fundamental recognition tasks, e.g., image classification \cite{krizhevsky2012imagenet} \cite{simonyan2014very} \cite{he2016deep} \cite{6020805}, image retrieval \cite{7006724} \cite{6882807} \cite{8019823}, object detection and recognition \cite{girshick2014rich} \cite{girshick2015fast} \cite{ren2015faster} \cite{tang2017object}. This requires the system to comprehensively understand the content of an image and bridge the gap between the image and the natural language. Automatically generating image descriptions is useful in multimedia retrieval, and image understanding.

Some pioneering research has been carried out in generating image descriptions \cite{kulkarni2013babytalk} \cite{fang2015captions}. However, as pointed out in \cite{karpathy2015deep}, most of these models often rely on hard-coded visual concepts and sentence templates, which limits their generalization capability. Recently, with the rapid development of deep learning in image recognition and natural language processing, the current trend of image captioning approaches \cite{vinyals2015show} is to follow the encoder-decoder framework, which shares the similarity with that in neural machine translation \cite{cho2014properties}. Most of these approaches represented the image as a single feature vector from the top layer of a pre-trained convolutional neural network (CNN) and cascaded recurrent neural network (RNN) to generate languages.

In fact, the tasks like image captioning and machine translation can be considered as a structured output problem where the task is to map the input to an output that possesses its own structure, as stated in \cite{cho2015describing}. An inherent challenge in these tasks is the structure of the output is closely related to the structure of the input. Hence, a key problem in these tasks is alignment \cite{cho2015describing}. Take neural machine translation for example, \cite{bahdanau2014neural} trained a neural model to softly align the output to the input for machine translation. Subsequent research \cite{xu2015show} applied the visual attention model to address this problem in image captioning, with much improvement. The visual attention mechanism is to dynamically select the relevant receptive fields in the CNN features to facilitate the image description generation, which, in other words, is to align the output words to spatial regions of the source image. In this paper, we also employ the visual attention mechanism for image captioning.

Nevertheless, natural language often consists of very meticulous descriptions, which correspond to the fine-grained objects of an image. As pointed out by \cite{8031355}, there are certain limitations of the most existing neural model-based schemes due to the mere use of the global feature representation in the image level. Some of the fine-grained objects might not to be recognized by only relying on the global image features. In this paper, we propose to use a pre-trained image detection model, i.e., Faster RCNN \cite{ren2015faster}, to retrieve the fine-grained image features from the top detected objects. These fine-grained object features, are able to provide complementary information for the global image representation, which will be proved in the experiments. In terms of the model structure, the object features are also processed by a visual attention mechanism, and are added to the original model to form a hierarchical feature representation and hence it is able to generate more meticulous descriptions.

In addition to the improvement of the image feature representation, we also consider to improve the current language model, which is widely used in neural machine translation and image captioning. An issue with most of the previous language model is the training framework, namely, the RNN using Maximum Likelihood Estimation (MLE) to generate image descriptions. As pointed out in \cite{bengio2015scheduled}, the MLE approaches suffer from the so-called exposure bias in the inference stage: the model generates a sequence iteratively and predicts the next token based on the previously predicted ones that may never be observed in the training data. In image description generation, the MLE also suffers from a problem that the generated languages do not correlate well with a human assessment of quality \cite{dai2017towards}.

Instead of only relying on the MLE, an alternative scheme is the generative adversarial network (GAN) \cite{goodfellow2014generative}. GAN was first proposed to generate realistic images. The GAN learns generative models without explicitly defining a loss function from the target distribution. Instead, GAN introduces a discriminator network which tries to differentiate real samples from generated samples. The whole network is trained using an adversarial training strategy. One can subsequently build a discriminator to judge how realistic are the samples generated by the description generator. The role of the caption generator, in this model, is similar with that of the the generator in the conditional GAN \cite{mirza2014conditional}, which is conditioned on the image features.

However, language generation is a discrete process. Directly providing the discrete samples as inputs to the discriminator does not allow the gradients to be back propagated through them. The reinforcement learning (RL) \cite{sutton1998reinforcement} framework provides a solution to estimate the gradients of the discontinuous units. The RL framework, when dealing with sequence generation, has the problem of lacking the intermediate reward, as discussed in \cite{yu2017seqgan}. The reward value can only be obtained when the whole sequence is generated. This is not suitable since what we want is the long-term reward of each intermediately generated token, so the whole sequence better optimized.

In the proposed scheme, the discriminator takes into account not only the differences between the generated captions and the reference captions but also the consistencies between captions and image features. Through the evaluation of the discriminator, the networks can better compensate for some unrealistic captions which might be generated under the MLE training. However, to deal with the discreteness of language, we treat the image captioning generator as an agent of RL. The feedbacks from the discriminator are considered as the rewards for the generator. To update the parameters of the image description generator in this framework, we consider the generator as a stochastic parameterized policy. We train the policy network using Policy Gradient \cite{sutton2000policy}, which naturally solve the differential difficulties in conventional GAN. Also, to solve the problem of lacking intermediate rewards, we borrow the idea from the famous ``AlphaGo" program \cite{silver2016mastering} in which a Monte Carlo roll-out strategy is applied to sample the expected long-term reward for an intermediate move. If we consider the sequence token generation as the the action to be taken in RL, we can apply a similar Monte Carlo roll-out strategy to obtain the intermediate rewards. \cite{yu2017seqgan} has successfully applied the Monte Carlo roll-out in sequence generation. In this paper, we use a similar sampling method to deal with intermediate rewards during the process of caption generation.

To summarize, our contribution in this paper is threefold:
\begin{itemize}
  \item We propose a hierarchical attention mechanism to reason on the global features and the local object features for image captioning.
  \item The policy gradient algorithm combined with the GAN is proposed for the training and optimization of the language model, with improvements over MLE training scheme.
  \item Through comprehensive experiments, we validate the proposed algorithm and comparable results with current state-of-the-art methods are achieved on the MSCOCO dataset.
\end{itemize}

\section{Related Work}

\subsection{Deep Model-based Image Captioning}

Promoted by the recent success of deep learning network in image recognition tasks and machine translation, the research on generating image description or image captioning has made remarkable progress \cite{mao2014deep} \cite{karpathy2015deep} \cite{fang2015captions} \cite{jia2015guiding} \cite{vinyals2015show} \cite{donahue2015long}. As mentioned above, most of the previously proposed approaches consider the image description generation as a translation process, mainly by borrowing the idea of the encoder-decoder framework \cite{cho2014learning} from neural machine translation \cite{cho2014properties}. Generally, this paradigm considers a deep CNN model as the image encoder, which maps the image into a static feature representation, and a RNN as a decoder to decode this static representations to an image description. The whole framework is trained using supervised learning under MLE. The generated description should be grammatically correct and match the content of the image.

Specifically, Karpathy et al. \cite{karpathy2015deep} proposed an alignment model through a multi-modal embedding layer. This model is able to align parts of a description with the corresponding regions of the image, which attracts significant attention. Jia et al. \cite{jia2015guiding} proposed a variation of LSTM, called gLSTM, for the image captioning task to mainly tackle the problem of losing track of the image content. This model includes the semantic information along with the whole image as inputs to generate captions. Donahue et al. \cite{donahue2015long} applied both of the convolutional layers and recurrent layers to form a Long-term Recurrent Convolutional Network (LRCN) for visual recognition and description.

Bahdanau et al. \cite{bahdanau2014neural} pointed out that a potential problem in this approach is that the model should compress all the necessary information of a source sentence into a fixed-length representation. This may make it difficult for the neural network to cope with long sentences. The static feature representation in the encoder-decoder framework, for both of machine translation and image captioning, cannot automatically retrieve relevant information from the source and thus at last influence the final performance. In neural machine translation, Bahdanau et al. \cite{bahdanau2014neural} proposed a kind of soft attention mechanism for machine translation, which enables the decoder to automatically focus on the relevant parts of the source sentence. In computer vision, the attention mechanism has long been the focus of much research \cite{bahdanau2014neural} \cite{mnih2014recurrent} \cite{ba2014multiple} since human perception does not tend to process a whole scene in its entirety at once but applies some mechanisms to selectively focus on the information needed. A comprehensive study for hard attention bound with reinforcement learning and soft attention for the task of image captioning was published by Xu et al. \cite{xu2015show}.

Yao et al. \cite{yao2015describing} tackled the video captioning task through capturing global temporal structures among video frames with a temporal attention mechanism, which makes the model dynamically focus on the key frames that are more relevant with the predicted word. Attention Models (ATT) developed by You et al. \cite{you2016image} first extracted semantic concept proposals and fused them with RNNs into hidden states and outputs. This method used K-NN, multi-label ranking to extract semantic concepts or attributes and fused these concepts into one vector using an attention mechanism. Similarly, Yao et al. \cite{8237786} embedded attributes with image features into a RNN with various methods to boost the image captioning performance. Recently, Chen et al. \cite{chen2017sca} proposed to combine the spatial attention and the channel-wise attention mechanism for image captioning, with improved results. Alternatively, Li et al. \cite{8031355} proposed a global-local attention mechanism to include local features extracted from the top detected objects from a pre-trained object detector. Inspired by \cite{8031355}, we also include the local features from top detected objects. However, we build a hierarchical model whilst they treated local and global features equivalently.

\subsection{Policy Gradient Optimization for Image Captioning}

Another approach to boost the performance of language tasks is to compensate the so-called exposure bias problem in RNN-based MLE learning. As pointed out in \cite{goyal2017actual}, RNNs are trained by MLE, which essentially minimized the KL-divergence between the distribution of target sequences and the distribution defined by the model. This KL-divergence objective tends to favour a model that overestimates its smoothness, which can lead to unrealistic samples \cite{goodfellow2016nips}.

In order to tackle the problems and generate more realistic image descriptions, some researches directly use evaluation metrics such as BLEU \cite{papineni2002bleu}, METEOR \cite{lavie2005meteor} and ROUGE \cite{lin2003automatic} as the reward signal and build the model under the RL framework. For instance, Ranzato et al. \cite{ranzato2015sequence} is the first research using the policy gradient algorithm in a RNN-based sequence model, in which a REINFORCE-based approach was used to calculate the sentence-level reward and a Monte-Carlo technique was employed for training. Liu et al. \cite{8237362} studied several linear combinations of the evaluation metrics and proposed to use a linear combination of SPICE \cite{anderson2016spice} and CIDEr \cite{vedantam2015cider} as the reward signal and apply a policy gradient algorithm to optimize the model, with improved results. This research used a Monte-Carlo roll-out strategy to obtain the intermediate reward during the process of description generation. More recently, Bahdanau et al. \cite{bahdanau2016actor}, instead of sentence-level reward in the training, applied the token-level reward in temporal difference training for sequence generation.

As discussed previously, the GAN \cite{goodfellow2014generative} estimates a difference measure using a binary classifier, called a discriminator, to discriminate between the target samples and generated samples. GANs rely on back-propagating these difference estimates through the generated samples to train the generator to minimize these differences. Hence, the whole network in GAN is trained in an adversarial way. The GAN was originally proposed to generate naturalist images \cite{goodfellow2014generative} \cite{mirza2014conditional} \cite{salimans2016improved} \cite{arjovsky2017wasserstein}. Directly applying a GAN for the language problem is impossible since sequences are composed of discrete elements in many application areas such as machine translation and image captioning.

A possible solution to tackle the discreteness problem of language is to use the Gumbel-Softmax approximation \cite{jang2016categorical} \cite{maddison2016concrete}. For instance, Shetty et al. \cite{shetty2017speaking} use a GAN to generate more realistic and accurate image descriptions with the aid of Gumbel-Softmax to deal with the discontinuousness issue in language processing. Another more general solution is to borrow an idea from the RL framework, in which the feedback from the discriminator is considered as the reward for the language generator. Dai et al. \cite{dai2017towards} built a model based on conditional GAN to generate diverse and naturalistic image descriptions and paragraphs, which utilizes a policy gradient for optimization. Yu et al. \cite{yu2017seqgan} proposed a model called SeqGAN, which unified the GAN framework and RL learning problem, this has recently received much attention \cite{kusner2016gans} \cite{wu2017adversarial}. They propose a three steps training strategy, which includes the pre-training the generator, pre-training the discriminator and the final adversarial training. In this paper, inspired by the SeqGAN, we propose to use a discriminator to judge the fitness of the generated image descriptions with reference to the image content and apply the policy gradient optimization technique \cite{sutton2000policy} to train the model. Unlike the original SeqGAN, our discriminator not only cares about the differences between the target language and model-generated language but also considers the coherence of the language with the image content.

\section{Approach}
In this section, we describe the proposed method based on two parts: the hierarchical attention mechanism and the policy gradient optimization algorithm.

\subsection{Hierarchical Attention Mechanism}\label{section}
The hierarchical attention mechanism consists of two parts: a spatial attention mechanism which corresponds to global CNN features and a local attention mechanism which corresponds to object features.

\begin{figure*}
  \centering
  \includegraphics[width=\linewidth]{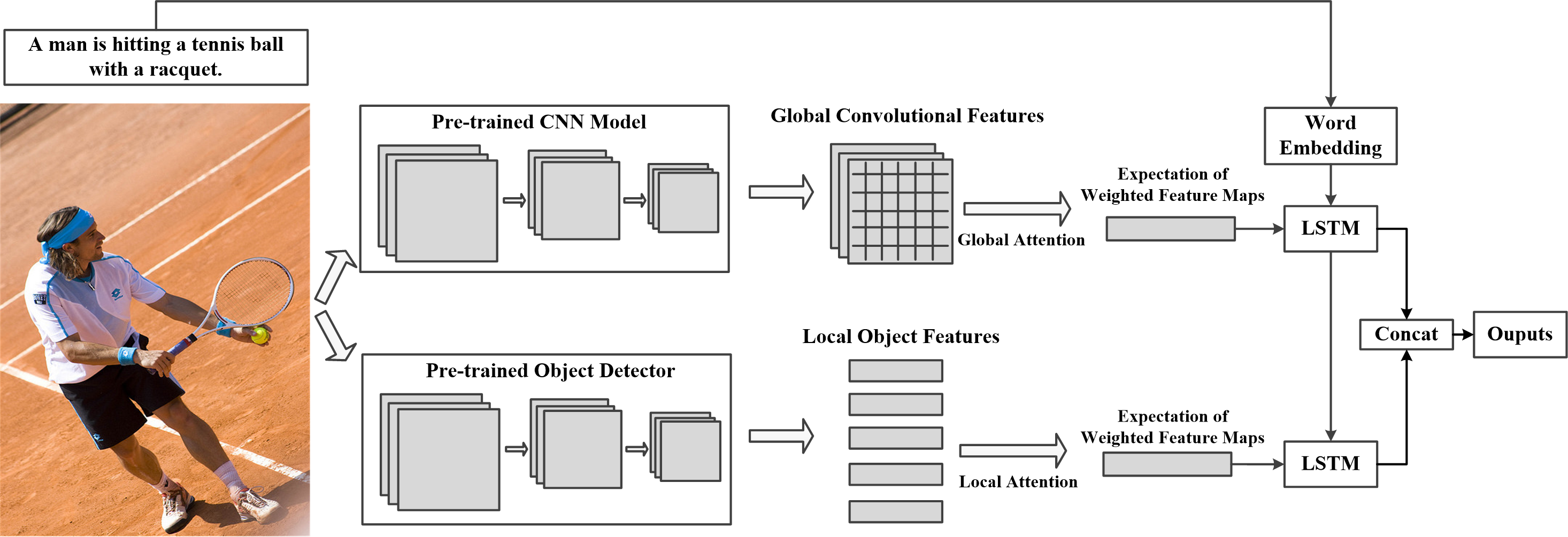}\\
  \caption{The hierarchical attention model structure: The CNN encoder and the object detector extracts the global and local features, respectively. These two types of features are forwarded to the LSTM models with the global and the local attention mechanisms. The outputs from the two LSTM models are concatenated and decoded to words. }\label{model}
\end{figure*}

The spatial attention mechanism is based on the model in \cite{xu2015show}. Specifically, the model comprises of an encoder and a decoder. We use a convolutional neural network pre-trained on the ImageNet dataset \cite{russakovsky2015imagenet} in order to extract a set of convolutional features. These features, denoted as $a = \{ a_1,...,a_L \}$, correspond to certain portions of the 2-D image. We extract convolutional features instead of fully connected ones in order to build a spatial attention mechanism since convolutional features have a spatial layout.

The Long-short Term Memory (LSTM) network, originally proposed by Hochreiter and Schmidhuber in \cite{hochreiter1997long}, is applied as the language decoder because of its superior performance in natural language processing.

\begin{equation}\label{LSTM}
\begin{aligned}
&  i_t = \sigma({W_{xi}}\ast{z_t}+{W_{hi}}\ast{h_{t-1}}+b_i) \\
&  f_t = \sigma({W_{xf}}\ast{z_t}+{W_{hf}}\ast{h_{t-1}}+b_f) \\
&  o_t = \sigma({W_{xo}}\ast{z_t}+{W_{ho}}\ast{h_{t-1}}+b_o) \\
&  g_t = \sigma({W_{xc}}\ast{z_t}+{W_{hc}}\ast{h_{t-1}}+b_c) \\
&  c_t = f_t \cdot c_{t-1} + i_t \cdot g_t \\
&  h_t = o_t \cdot \phi(c_t)
  \end{aligned}
\end{equation}

In Equation \ref{LSTM}, $i_t$, $f_t$, $o_t$, $c_t$ and $h_t$ are the input gate, forget gate, output gate, cell memory and hidden state of a LSTM network, respectively. $g_t$ and $h_t$ are the input and the output of the LSTM model. $z_t$ is the context vector, which can be processed by the soft attention mechanism and is able to capture visual information associated with a certain input location. The soft attention mechanism has to automatically allocate adaptive weights for the image locations to facilitate the task at hand.

\begin{equation}\label{weights}
  e_{ti} = f_{att} (a_i, h_{t-1})
\end{equation}
where $a_i \in \{ a_1,...,a_L \}$. Equation \ref{weights} actually maps the image features from each location, along with information from the hidden state, into an adaptive weight, which indicates the importance of each image location for the recognition.

\begin{equation}\label{softmax}
 \alpha_{ti} = \frac{exp(e_{ti})}{\sum_{k=1}^{L} exp(e_{tk})}
  \end{equation}

Then, Equation \ref {softmax} normalizes the adaptive weights into a probability value in the range of 0 and 1 using the Softmax function. Once these weights (summed to 1) are computed, we element-wisely multiply the weights vector $\alpha_t$ with image feature vector $a$ and sum them to the context vector $z_t$, which can be expressed as in Equation \ref {context}. This can be seen as the expectation of weighted features maps.

\begin{equation}\label{context}
 z_t = \sum_{i=1}^L \alpha_{t,i}a_{i}
   \end{equation}

Then the context vector $z_t$ is forwarded to the LSTM network to generate captions, as described in Equation \ref {LSTM}. This soft attention mechanism is able to adaptively select the relevant visual parts of the given image features and thus facilitate the recognition.

The local attention mechanism is formulated using object features and another LSTM model. We use a pre-trained object detector to retrieve the top $N$ detected object features, which are denoted as $d = \{ d_1,...,d_N \}$. We then use another LSTM model with soft attention to allocate adaptive weights to each of these features.

\begin{equation}\label{weights_detector}
  e^{d}_{ti} = f^{d}_{att} (d_i, h^{d}_{t-1})
\end{equation}
where $h^d$ indicates the hidden state of the LSTM model for the local attention mechanism.

\begin{equation}\label{softmax_detector}
 \alpha^{d}_{ti} = \frac{exp(e^{d}_{ti})}{\sum_{k=1}^{L} exp(e^{d}_{tk})}
  \end{equation}
Similarly, Equation \ref{softmax_detector} normalizes the adaptive weights for local features to a probability value with the Softmax function.

\begin{equation}\label{context_detector}
 z^{d}_t = Concat(\sum_{i=1}^N \alpha^{d}_{t,i} d_{i}, h_{t-1})
   \end{equation}
Equation \ref{context_detector} demonstrates that the context vector for local attention model catching information from both the local features and the global attention mechanism, where $Concat$ indicates the concatenation operation of the features. This context vector is then forwarded to a second LSTM model as described by Equation \ref{LSTM2}.

\begin{equation}\label{LSTM2}
\begin{aligned}
&  i^{d}_t = \sigma({W^{d}_{xi}}\ast{z^{d}_t}+{W^{d}_{hi}}\ast{h^{d}_{t-1}}+b^{d}_i) \\
&  f^{d}_t = \sigma({W^{d}_{xf}}\ast{z^{d}_t}+{W^{d}_{hf}}\ast{h^{d}_{t-1}}+b^{d}_f) \\
&  o^{d}_t = \sigma({W^{d}_{xo}}\ast{z^{d}_t}+{W^{d}_{ho}}\ast{h^{d}_{t-1}}+b^{d}_o) \\
&  g^{d}_t = \sigma({W^{d}_{xc}}\ast{z^{d}_t}+{W^{d}_{hc}}\ast{h^{d}_{t-1}}+b^{d}_c) \\
&  c^{d}_t = f^{d}_t \cdot c^{d}_{t-1} + i^{d}_t \cdot g^{d}_t \\
&  h^{d}_t = o^{d}_t \cdot \phi(c^{d}_t)
  \end{aligned}
\end{equation}

The two LSTM models, denoted as $LSTM^{G}$ for the global features and $LSTM^{L}$ for the local features are jointly trained to map the hierarchical feature representation with language. $LSTM^{L}$ is at a higher level, which can be used to decode the hidden states for the final outputs. However, the gradient vanishing problem cannot be avoided if we only use the hidden states from $LSTM^{L}$ to decode information. Inspired by \cite{he2016deep} in which a shortcut in network connections is applied to solve the gradient vanishing problem, we concatenate the hidden states from $LSTM^{G}$ and $LSTM^{L}$ to decode and map the hidden states to language vectors, which can be seen in Equation \ref{decode}.
\begin{equation}\label{decode}
\begin{aligned}
& h^{output}_t = Concat(h_t, h^{d}_t) \\
& logits = W_p h^{output}_t \\
& P(s_t|I, s_0, s_1, s_2, ..., s_{t-1}) = Softmax(logits)
   \end{aligned}
   \end{equation}

In MLE training, if the length of a sentence is $T$, the loss function can be formulated as in Equation \ref{loss}, which is the sum of the log likelihood of each word.

\begin{equation}\label{loss}
Loss =  \sum_{i=0}^{T} log(p(s_t|I, s_0, s_1, s_2, ..., s_{i}))
   \end{equation}

\subsection{Policy Gradient Optimization}
In addition to only using the MLE to train the image caption generator, to alleviate the previously discussed exposure bias problem in RNN-based MLE training as discussed previously, we also apply a policy gradient optimization algorithm in the RL framework to increase the quality of the generated descriptions.
\begin{figure*}
  \centering
  \includegraphics[width=\linewidth]{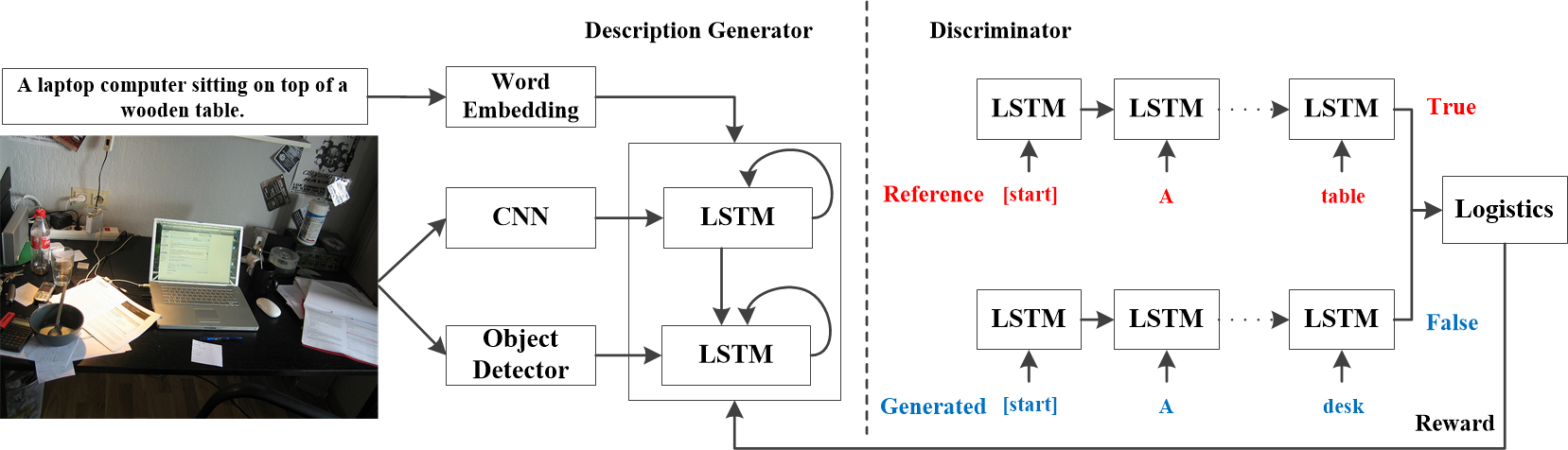}\\
  \caption{Policy Gradient optimization with a discriminator to evaluate the similarity between the generated sentence and the reference sentence. }\label{gan_word}
\end{figure*}

\begin{figure*}
  \centering
  \includegraphics[width=\linewidth]{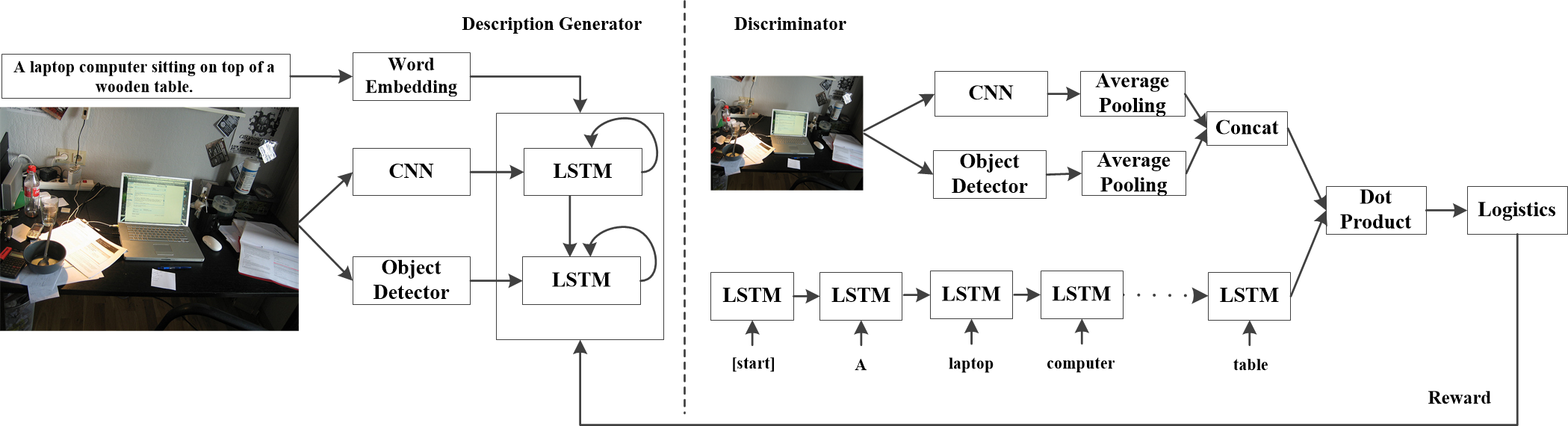}\\
  \caption{Policy Gradient optimization with a discriminator to evaluate the coherence between the generated sentence and the image contents.}\label{gan}
\end{figure*}

We feed both of the generated descriptions and the reference descriptions to the discriminator. The level of coherence of the descriptions and image content is calculated by the dot product, which is forwarded to the discriminator, as described in Fig. \ref{gan}. This operation is to consider the coherence between certain captions (sequences) and corresponding image features, which is able to make the generated captions more realistic and naturalistic. The reference sequences are labeled as true whilst the generated sequences are labeled as false during the training of the discriminator. The model is also a LSTM network with Softmax Cross Entropy loss. Hence, the discriminator outputs the probabilities of a sample being true. These probabilities, are then considered as the reward signal in the RL framework, to be utilized in the Policy Gradient algorithm for updating the parameters of the image caption generator.

Following \cite{sutton2000policy}, the objective of the policy network $G_{\theta}(y_t|y_{1:t-1})$ (the image caption generator), is to generate a sequence from the start state $S_0$ to maximize its expected long-term reward as described by Equation \ref{long-term}:

\begin{equation}\label{long-term}
J(\theta)= E[R_T|s_0, \theta] = \sum_{y_1{\in}Y}G_{\theta}(y_1|s_0) \cdot Q_{D_{\theta}}^{G_{\theta}}(s_0,y_1)
   \end{equation}
where $R_T$ is the reward for a complete sequence. $Q_{D_{\theta}}^{G_{\theta}}(s,y)$ is the action-value function of a language sequence, which is defined as the expected accumulative reward starting from state $s$, taking a certain action, and then following policy $G_{\theta}$.

The action-value function is estimated using the REINFORCE algorithm \cite{williams1992simple} and considers the probability of being real generated by the discriminator as a reward, which can be defined as in Equation \ref{d-reward}.

\begin{equation}\label{d-reward}
Q_{D_{\theta}}^{G_{\theta}}(a = y_T, s = Y_{1:T-1}) = D_{\theta}(Y_{1:T})
\end{equation}

As can be seen in Equation \ref{d-reward}, the discriminator only provides a reward for a complete sequence. We should not only care about the reward for a complete tokens but also the long-term reward for the future time-steps since the long-term reward is what we actually want. Similar to the game of Go \cite{silver2016mastering} in which the agent sometimes give up an immediate interest but cares about the final victory, we apply a similar Monte Carlo roll-out strategy for an intermediate state, i.e., an unfinished sequence. We represent an N-time Monte Carlo search as in Equation \ref{mc-rollout}.
\begin{equation}\label{mc-rollout}
\begin{aligned}
& {Y_{t+1:T}^1, ..., Y_{t+1:T}^n, ..., Y_{t+1:T}^N} = MC^{G_{\theta}}(Y_{1:t};N) \\
& MC  =  \sim Multinomial({logits})
\end{aligned}
\end{equation}

where $Y_{1:t}$ is the generated sequence tokens and $Y_{t+1:T}^n$ is the Monte Carlo sampled based on a roll-out policy, which, in our case, is set as the same as the image caption generator for convenience. In reality, we can use any policy to perform the roll-out operation. $logits$ is the output of the LSTM decoder. MC is defined as a sampling procedure from a Multinomial distribution.

If there is no intermediate reward, the Monte Carlo roll-out strategy can sample the future possible tokens $N$ times and average these rewards to achieve the goal of reward estimation, which is described in Equation \ref{mc-rollout2}.
\begin{equation}\label{mc-rollout2}
\begin{aligned}
& Q_{D_{\theta}}^{G_{\theta}}(a = y_t, s = Y_{1:t-1}) =  \\
    & \frac{1}{N}\sum_{n=1}^{N}D_\theta(Y_{1:T}^n), Y_{1:T}^n\in{MC^{G_{\theta}}(Y_{1:t};N)}, & \ for \  t<T \\
    &  D_\theta(Y_{1:T}), & \ for \  t=T
\end{aligned}
\end{equation}

\begin{figure}[!ht]
  \begin{center}
  \includegraphics[width=\linewidth]{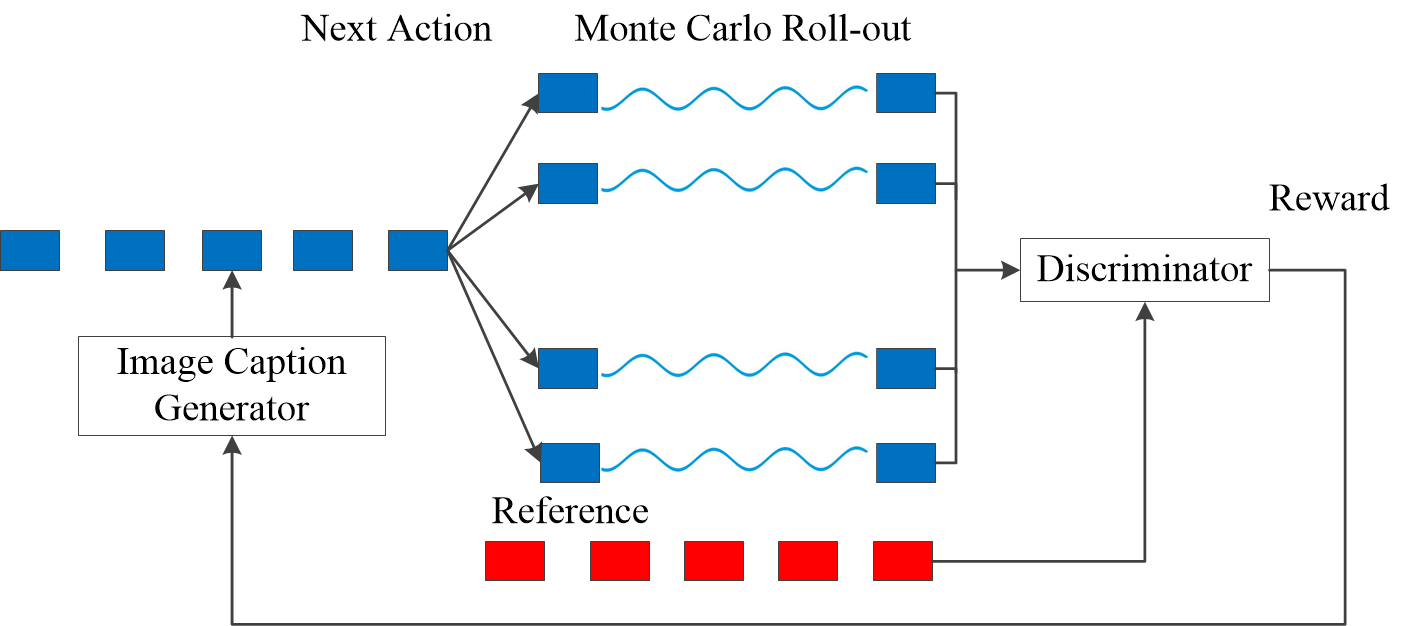}\\
  \caption{Monte Carlo roll-out: We use Monte Carlo sampling to sample tokens in the future time steps and average them to obtain the intermediate rewards so as to optimize the token generated at each time step.}\label{montecarlo}
  \end{center}
\end{figure}

The Monte Carlo roll-out strategy can be better visualized in Fig. \ref{montecarlo}.

Once the reward value from the discriminator is obtained, it is ready to update the generator. The goal is to maximize the average reward starting from the initial state as defined in Equation \ref{av-reward}.

\begin{equation}\label{av-reward}
J(\theta) = \frac{1}{N}\sum_{i=1}^{N}{V_\theta(s_0|X_i, Y_i)}
\end{equation}
where $N$ is the number of samples used for training. We can use the Policy Gradient theorem from \cite{sutton2000policy} and write the gradient of the objective function (reward signal) as in Equation \ref{gradient}.
\begin{equation}\label{gradient}
\bigtriangledown_{\theta}J(\theta) = E_{Y_{1:t-1}{\sim}G_{\theta}}[\sum_{y_t{\in}Y}\bigtriangledown{G_{\theta}(y_t|Y_{1:t-1})\cdot{Q_{D_{\theta}}^{G_{\theta}}(Y_{1:t-1}, y_t)}}]
\end{equation}
Since the expectation can be approximated by sampling, we can now update the parameters of the image caption generator using Equation \ref{update}.
\begin{equation}\label{update}
\theta \gets \theta + \alpha_h\bigtriangledown_{\theta}J(\theta)
\end{equation}
In practice, we can use advanced gradient algorithms such as RMSprop \cite{tieleman2012lecture} and Adam \cite{kingma2014adam} in training the caption generator.

The image caption generator and discriminator are adversarially trained in the framework of GAN \cite{goodfellow2014generative}. In GAN \cite{mirza2014conditional}, the discriminator can pass the gradient directly to the generator. Due to the discreteness of the sequence generation, we apply RL to estimate the gradient of the generator in our model.

Specifically, the training strategy is described in Algorithm \ref{Algorithm}. We initially pre-train the image caption generator using MLE. In practice, this is equivalent to the Cross Entropy loss \cite{de2005tutorial}. Hence, we can set the pre-training step the same as in \cite{xu2015show}. The trained model is used to generate some captions which are set as fake samples, which, along with the reference captions, are fed into the discriminator for training. Similarly, the discriminator is also pre-trained for certain steps. The next steps are the adversarial training steps, in which the image caption generator and discriminator are trained alternatively until convergence of the networks.

In addition to the sentence comparison scheme introduced previously, and shown in Fig. \ref{gan_word}, we also employ a scheme to evaluate the coherence between the generated captions and the image content. Specifically, both of the global features and local object features are processed by average pooling in order to obtain fixed-size feature representation, denoted as $V_i$. The captions, similar to the sentence comparison scheme, are also encoded into a fixed-size vector, using a LSTM model, denoted as $V_w$. The two vectors $V_i$ and $V_w$ are then dot producted and forwarded to logistic function to obtain the reward for RL training, which can be seen in Fig. \ref{gan}.

\begin{algorithm}
\caption{Image Caption Generation by Adversarial Training and Reinforcement Learning}
\label{Algorithm}
\begin{algorithmic}
\REQUIRE Image Caption Generator $G_{\theta}$; Discriminator $D_{\theta}$.
\STATE Pre-training $G_{\theta}$ using MLE by some epoches.
\STATE Generating negative samples using pre-trained $G_{\theta}$ to train $D_{\theta}$.
\STATE Pre-training $D_{\theta}$ by some steps.
\REPEAT
\FOR{update-generator for 1 step}
\STATE Generate a sequence $Y_{1:T} = (y_1, .., y_T)$.
\FOR {$t=1$ to $T$}
\STATE Compute the intermediate reward $Q(t)$ by Monte Carlo roll-out.
\ENDFOR
\STATE Update the parameters $\theta$ using Policy Gradient.
\ENDFOR
\FOR{update-discriminator for 1 step or 5 steps}
\STATE Training discriminator $D_{\theta}$ using reference sequence (True) and generated sequence (Fake) using current generator.
\ENDFOR
\UNTIL Convergence
\end{algorithmic}
\end{algorithm}

\section{Experimental Validation}

\subsection{Dataset Introduction}

We conduct our experiments using the MSCOCO dataset \cite{lin2014microsoft}. To be consistent with the previous researches, we use the MSCOCO 2014 released version, which includes 123,000 images. The dataset contains 82,783 images in the training set, 40,504 images in the validation set and 40,775 images in the test set. As the ground-truth for the MSCOCO test set is not available, the validation set is further splited into a validation subset for model selection and a test subset for local experiments. This is the ``Karpathy'' split \cite{karpathy2015deep}. It utilizes the whole 82,783 training set images for training, and selects 5,000 images for validation and 5,000 images for testing from the official validation set. The standard evaluation protocol contains BLEU \cite{papineni2002bleu},  METEOR \cite{lavie2005meteor}, CIDEr \cite{vedantam2015cider} and ROUGE-L \cite{lin2003automatic}.

BLEU is the most popular metric for the performance evaluation in machine translation. The metric is only based on the n-gram statistics. The BLEU-1, BLEU-2, BLEU-3 and BLEU-4 measure the performance of the 1, 2, 3, 4-gram, respectively. METEOR is based on the harmonic mean of unigram precision and recall, and seeks correlation at the corpus level. CIDEr can be used to evaluate the generated sentences with human consensus. ROUGE-L measures the common maximum-length subsequence for the target sentence and the generated sentence.

\subsection{Implementation Details}
For all the images in the COCO dataset, we obtain global convolutional features (from the layer ``res5c'') using a pre-trained Residual-152 network \cite{he2016deep} on the platform of Caffe \cite{jia2014caffe}, with a dimensionality of $49\times2048$. We also retrieve local object features using a Faster RCNN \cite{ren2015faster} object detection network pre-trained on the MSCOCO dataset. Specifically, we obtain the top $K$ detected object features from the layer of ``FC6'' layer of the VGG16 model \cite{simonyan2014very} used in Faster RCNN, with dimensionality of $K\times4096$.  We build the hierarchical attention mechanism and policy gradient optimization on the TensorFlow platform \cite{abadi2016tensorflow}.

\subsubsection{Training the Faster RCNN on the MSCOCO dataset}
In order to obtain better local object features, we train the Faster RCNN model on MSCOCO object detection dataset. The model is first pre-trained on the ImageNet object detection dataset \cite{russakovsky2015imagenet}. The MSCOCO object detection dataset shares the same images with the image caption task. Consequently, we keep the same splits with the image caption dataset for training. The training process on the MSCOCO dataset is almost the same with the pre-training on ImageNet. The initial learning rate is set to 0.001. The momentum of the stochastic gradient descent is set to 0.9 and the weight decay is set to 0.0005.

\subsubsection{Language Pre-processing}
To pre-process the language, the special symbols such as `.', `,', `(', `)' and `-' are replaced with blank spaces whilst `\&' is replaced with `and'. Since we set the maximum length of the descriptions as 20 words, we delete the caption references from the original dataset which are longer than 20. For the vocabulary establishment, following the open-source code of \cite{karpathy2015deep}, we include words that occurs more than 5 times in the vocabulary. We map the symbol `NULL' to 0, `START' to 1 and `END' to 2.

\subsubsection{Training Details of the Model}
The network was first pre-trained using MLE for 10 epochs. During training, the size of the hidden states of the two LSTM models is set as 512. We choose the same size of hidden states of \cite{8031355} as they achieved satisfactory performance with this size of the hidden states. We set the batch size as 32 and the learning rate as 0.001 and we use the Adam algorithm \cite{kingma2014adam} to train the network. Subsequently, we train the discriminator for 2500 steps, following by an adversarial training scheme, in which the caption generator and discriminator are trained alternatively until convergence. During the pre-training steps of the discriminator and the policy gradient-based adversarial training as described previously, the Adam algorithm is also applied. The learning rate for these steps are set as 0.0001. Following the open-source code of \cite{karpathy2015deep}, at training time, we set the maximum length of the input sequence to 20 words. During the testing time, alternatively, we set maximum length of a generated symbols as 30 words. During the training of the proposed model, we add a trainable word embedding layer from Google's TensorFlow platform \cite{abadi2016tensorflow}. All the experiments are conducted on a server embedded with NVIDIA TITAN X GPU and installed with the Ubuntu 14.04 operating system.

\subsection{Results}

\begin{figure*}
  \centering
  \includegraphics[width=\linewidth]{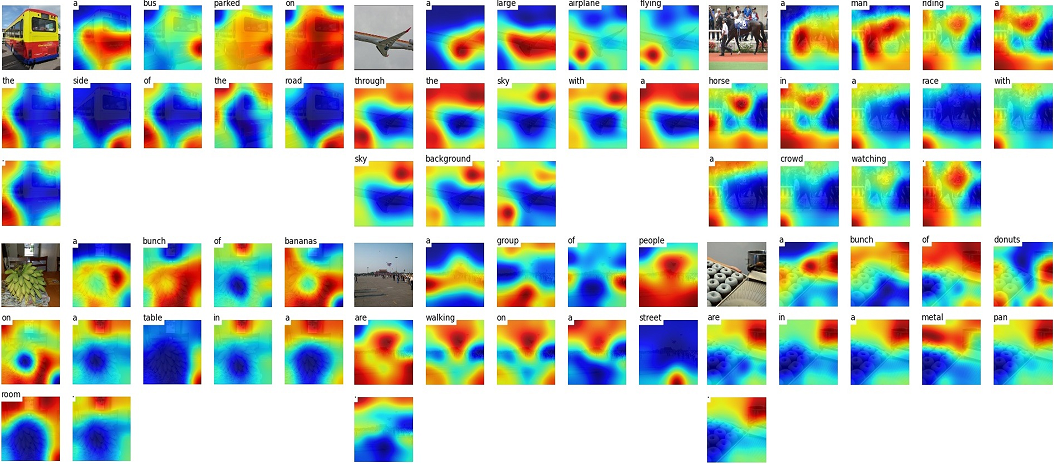}\\
  \caption{Visualization of the global attention maps and generated captions. The red color indicates the importance of each region of the image.}\label{vis_global}
\end{figure*}

\begin{figure}
  \centering
  \includegraphics[width=\linewidth]{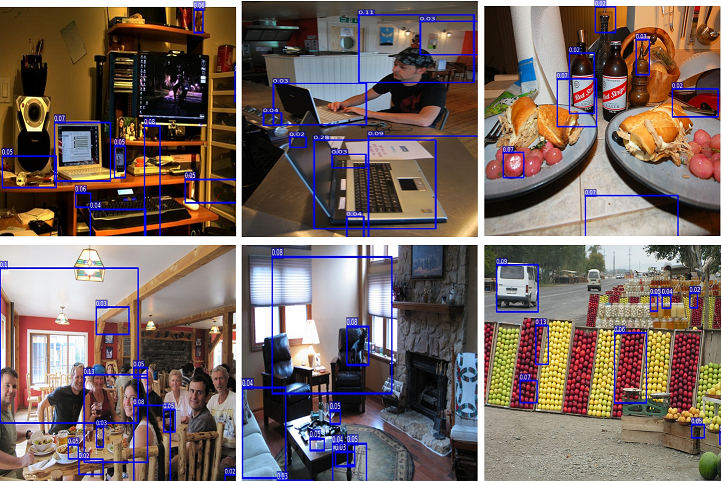}\\
  \caption{Visualization of the attentive weights on the top 10 detected objects, the blue boxes indicate the detected objects whilst the labels show the attentive weights of the local attention model. }\label{vis_local}
\end{figure}

\begin{figure*}[!t]
\centering
  \subfigure[\newline Ground-truth:  \newline \textcolor{green}{A group of people standing next to a bus under an airplane .}  \newline MLE:  \newline  \textcolor{blue}{A large airplane is parked on the runway.}  \newline Ours:  \newline  \textcolor{red}{A large airplane is parked on the runway with people walking around.}]{
  \includegraphics[width =0.2\textwidth]{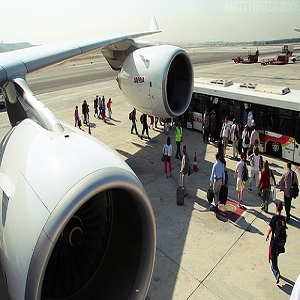}}
  \subfigure[\newline Ground-truth:  \newline \textcolor{green}{A yellow and red bus parked in a parking lot with other busses.}  \newline MLE:  \newline  \textcolor{blue}{A yellow bus is parked on the side of the road.}  \newline Ours:  \newline  \textcolor{red}{A yellow and red bus parked in a parking lot.}]{
  \includegraphics[width =0.2\textwidth]{pics/img524992}}
  \subfigure[\newline Ground-truth:  \newline \textcolor{green}{A little boy sitting in front of a hot dog covered in ketchup.}  \newline MLE:  \newline  \textcolor{blue}{A little girl is eating a hot dog.}  \newline Ours:  \newline  \textcolor{red}{A young boy is eating a hot dog.}]{
  \includegraphics[width =0.2\textwidth]{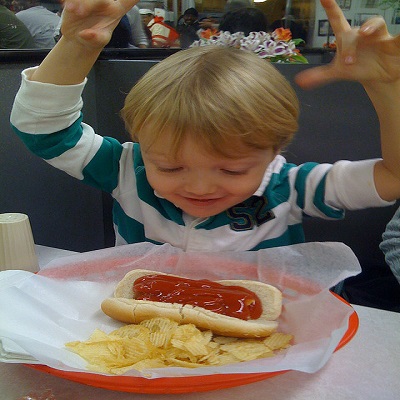}}
  \subfigure[\newline Ground-truth:  \newline \textcolor{green}{The lone adult cow walks on rocks near the beach.}  \newline MLE:  \newline  \textcolor{blue}{A cow is walking down the street in the sand.} \newline Ours: \newline \textcolor{red} {A cow is standing on the beach next to body of water.}]{
  \includegraphics[width =0.2\textwidth]{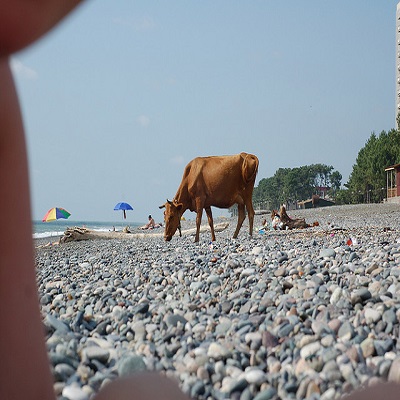}}
   \subfigure[\newline Ground-truth:  \newline \textcolor{green}{A baseball player swinging a baseball bat during a game.}  \newline MLE:  \newline  \textcolor{blue}{A baseball player is preparing to swing at a pitch.} \newline  Ours: \newline \textcolor{red}{A baseball player is swinging a bat at a ball.}]{
  \includegraphics[width =0.2\textwidth]{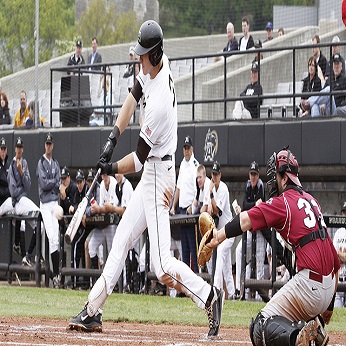}}
 \subfigure[\newline Ground-truth:  \newline \textcolor{green}{Six cows standing and laying on the beach.}  \newline MLE:  \newline  \textcolor{blue}{A group of cows standing on top of a snow covered field.} \newline  Ours: \newline \textcolor{red}{A group of cows standing on top of a sandy beach.}]{
  \includegraphics[width =0.2\textwidth]{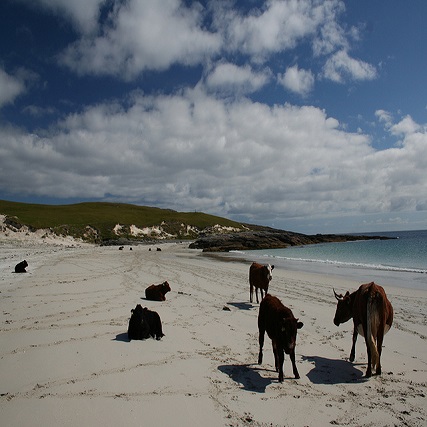}}
  \subfigure[\newline Ground-truth:  \newline \textcolor{green}{A fat cat in the living room watching the tv.}  \newline MLE:  \newline  \textcolor{blue}{A cat is sitting in a living room with a television.} \newline  Ours: \newline \textcolor{red}{A cat sitting on the floor watching a television.}]{
  \includegraphics[width =0.2\textwidth]{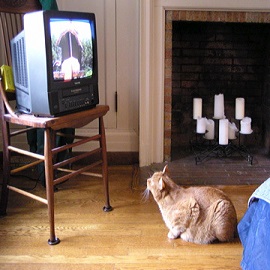}}
\subfigure[\newline Ground-truth:  \newline \textcolor{green}{A giraffe is walking through the forest with tall trees.}  \newline MLE:  \newline  \textcolor{blue}{A giraffe is standing in the woods with trees in the background.} \newline  Ours: \newline \textcolor{red}{A giraffe standing next to a tree in a forest.}]{
  \includegraphics[width =0.2\textwidth]{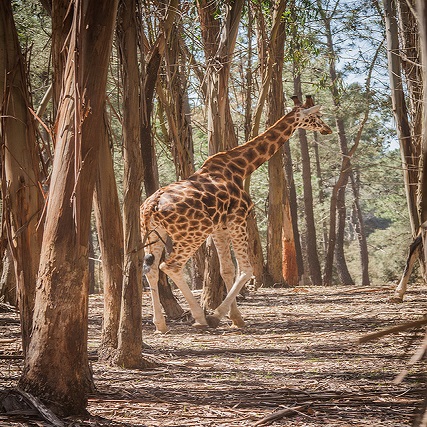}}
  \caption{Visualization of the generated descriptions: the red color texts indicate the captions generated by our model, which is more accurate and realistic than blue text captions generated by MLE model. All samples are randomly selected.}
  \label{vis_caption}
\end{figure*}

\subsubsection{Quantitative Evaluation}

In this section, a comprehensive quantitative evaluation is conducted using different experimental settings on the MSCOCO dataset.

\paragraph{Comparison between the global attention, the local attention and the hierarchical attention model}

We first obtain the results using only the global attention model, which is similar to the soft attention model in \cite{xu2015show}. Since we use advanced CNN features from the Residual-152 model, the results of BLEU, METEOR, CIDEr and ROUGE-L are all satisfactory, and are listed in Table \ref{Tab:spatial}. Then only the local attention model using the detected object features from a Faster RCNN detector is tested, with results which are much lower than those for the global attention model as listed in Table \ref{Tab:spatial}. One of the possible reasons is that the Faster RCNN only uses the VGG16 model, which is not as powerful as the Residual-152 network. Another reason is that the local object features, despite the capability to provide complementary information to the global attention model, can sometimes miss many important features. Finally, we test our proposed hierarchical attention model under MLE training, which utilizes both of the global and local attention for image captioning. The results improve the baseline significantly, which can be seen in Table \ref{Tab:spatial}. Specifically, all of the seven evaluation metrics are improved using our hierarchical attention model.

\paragraph{The determination of the number of top detected objects}

To determine the best number $k$ for the top detected objects in the local attention model, we perform an ablation study. We extract the 10, 20 and 30 top detected object features and test them using the hierarchical attention model. The results can be seen in Table \ref{Tab:local}. With the increase of the number $k$ from 10 to 30, the performance increases accordingly. Although the maximum length of our generated sentences is set as 30, not every word represents an object. Also, intuitively, there are a maximum 30 objects within an image. Hence, in the following experiments, we use the 30 top detected object features for the local attention model.

\begin{table*}[!t]
\caption{Comparison of image captioning using different attention mechanism results on the MSCOCO dataset}
 \centering
\begin{tabular}{|c|c|c|c|c|c|c|c|}
  \hline
  \hline
  Methods  & BLEU-1 & BLEU-2 & BLEU-3 & BLEU-4 & METEOR & CIDEr & ROUGE-L \\
  \hline
  Soft Attention \cite{xu2015show} & 70.7 & 49.2 & 34.4 & 24.3 & 23.90 & - & - \\
  \hline
  Global Attention & 70.121 & 50.304 & 35.434 & 25.111 & 23.658 & 84.701 & 54.308  \\
  Local Attention & 64.059 & 42.359 & 28.089 & 19.033 & 20.203 & 56.898 & 49.861 \\
  Hierarchical Attention & \textbf{72.611} & \textbf{52.769} & \textbf{37.802} & \textbf{27.243} & \textbf{24.731} & \textbf{88.140} & \textbf{56.048} \\
  \hline
  \hline
\end{tabular}
\label{Tab:spatial}
\end{table*}%

\begin{table*}[!t]
\caption{Comparison of image captioning results on the MSCOCO dataset with different numbers of objects}
 \centering
\begin{tabular}{|c|c|c|c|c|c|c|c|}
  \hline
  \hline
  Methods  & BLEU-1 & BLEU-2 & BLEU-3 & BLEU-4 & METEOR & CIDEr & ROUGE-L \\
  \hline
   Hierarchical Attention with 10 Objects for Local Attention & 70.601 & 50.423 & 36.643 & 25.389 & 24.633 & 87.316 & 55.241 \\
   Hierarchical Attention with 20 Objects for Local Attention & 72.159 & 52.498 & 37.552 & 26.918 & 24.725 & \textbf{88.639} & 55.825 \\
   Hierarchical Attention with 30 Objects for Local Attention & \textbf{72.611} & \textbf{52.769} & \textbf{37.802} & \textbf{27.243} & \textbf{24.731} & 88.140 & \textbf{56.048} \\
  \hline
  \hline
\end{tabular}
\label{Tab:local}
\end{table*}%

\begin{table*}[!t]
\caption{Comparison of image captioning results on the MSCOCO dataset with different settings for policy gradient (PG) optimization}
 \centering
\begin{tabular}{|c|c|c|c|c|c|c|c|}
  \hline
  \hline
  Methods  & BLEU-1 & BLEU-2 & BLEU-3 & BLEU-4 & METEOR & CIDEr & ROUGE-L \\
  \hline
  MLE training only & \textbf{72.611} & 52.769 & 37.802 & 27.243 & 24.731 & 88.140 & 56.048 \\
  \hline
  PG with 2500 steps for pre-training D followed by 1 D and 1 G step & 72.450 & \textbf{52.845} & \textbf{38.141} & 27.551 & 24.543 & 87.416 & \textbf{55.876} \\
  PG with 2500 steps for pre-training D followed by 5 D and 1 G step & 72.104 & 52.739 & 38.122 & \textbf{27.602} & \textbf{24.928} & \textbf{89.072} & 56.063 \\
  \hline
  \hline
\end{tabular}
\label{Tab:gan_word}
\end{table*}%

\paragraph{The performance of Policy Gradient with reward only from language comparison}

\begin{figure}
  \centering
  \includegraphics[width=\linewidth]{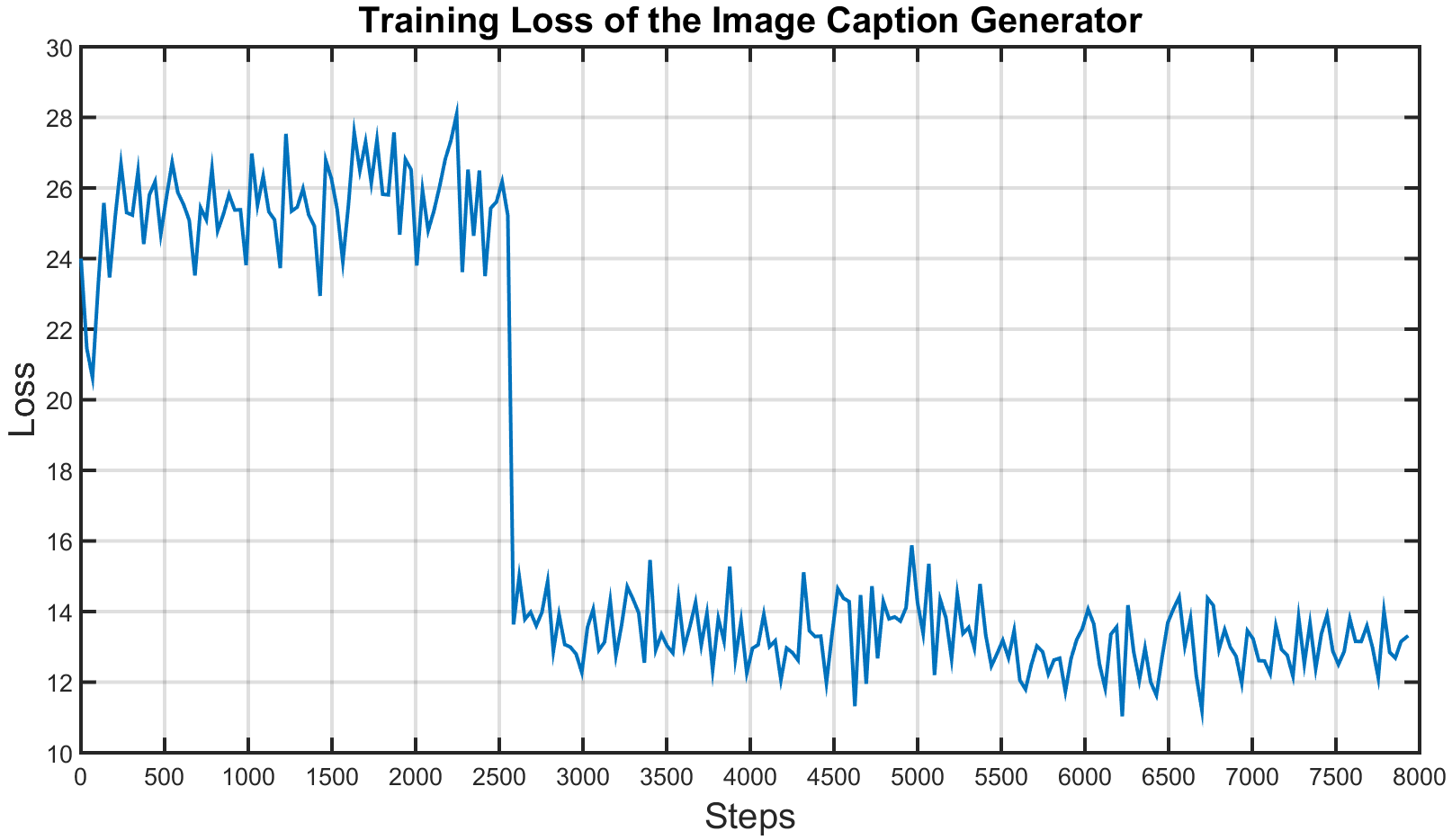}\\
  \caption{The loss curve of the image caption generator during reinforcement learning steps: before 2500 iterations, we pre-train the discriminator. Starting from the 2500 iterations, we start the adversarial training of the generator and discriminator. The loss value starts to decrease starting from 2500 iterations as the parameters of the generator begins to be updated. }\label{gen}
\end{figure}

Next we start the reinforcement learning steps. We first train the discriminator which only compares the similarity between the reference sentence and the generated sentence. Specifically, we follow the model defined in Fig. \ref{gan_word}. The discriminator is first trained in 2500 steps, which we find sufficient for the discriminator to converge. The loss curve of the image caption generator is shown in Fig. \ref{gen}. After 2500 steps pre-training the discriminator, the loss of the image caption generator starts to decline, which validates that the policy gradient starts to work. Then we further train the generator and discriminator adversarially for another 1 epoch, and report the results in Table \ref{Tab:gan_word}. We also experimented with two different settings in the adversarial training steps. The first setting is to train 1 step for the discriminator, followed by another step for the generator. Another setting is to train the discriminator for 5 steps, followed by 1 step training for the generator. We find the final results of the two setting are similar, which all slightly improve the MLE training baseline. The reason for the improvement is because the reinforcement learning solves the exposure bias problem during MLE training. However, this scheme lacks the measurement of the similarity between the generated descriptions and the image contents, which prevents the image caption generator from generating more naturalistic and diverse descriptions.

%

\begin{table*}[!t]
\caption{Comparison of image captioning results on the MSCOCO dataset for policy gradient (PG) optimization with discriminator for evaluation of the coherence between language and image content. }
 \centering
\begin{tabular}{|c|c|c|c|c|c|c|c|}
  \hline
  \hline
  Methods  & BLEU-1 & BLEU-2 & BLEU-3 & BLEU-4 & METEOR & CIDEr & ROUGE-L \\
  \hline
  MLE training only & 72.611 & 52.769 & 37.802 & 27.243 & 24.731 & 88.140 & 56.048 \\
  \hline
  Global Attention & 70.121 & 50.304 & 35.434 & 25.111 & 23.658 & 84.701 & 54.308  \\
  PG with similarity of global features (1 D and 1 G step) & 72.250 & 52.290 & 37.099 & 26.331 & 23.815 & 84.516 & 55.238 \\
  PG with similarity of global features (5 D and 1 G step) & 72.234 & 52.120 & 36.887 & 26.065 & 23.957  & 84.224 & 55.244\\
  \hline
  PG with similarity of global-local features (1 D and 1 G step) & \textbf{73.036} & \textbf{53.688} & \textbf{39.069} & \textbf{28.551} & \textbf{25.324} & \textbf{92.449} & \textbf{56.539} \\

  \hline
  \hline
\end{tabular}
\label{Tab:gan}
\end{table*}%


\paragraph{The performance of Policy Gradient with reward from the measurement of coherence between language and image content}

To train the image caption generator to generate more naturalistic and diverse descriptions, we further test the model defined in Fig. \ref{gan_word}. First we only extract the global features and perform average pooling, resulting with a feature dimension of 2048. We then use the dot product to measure these image features and language embedding features by a discriminator, which can be considered as the reward within the reinforcement learning framework. The experimental results from this model can be seen in Table \ref{Tab:gan}.

However, the results from all of the seven metrics are even lower than the MLE training baseline. One possible reason, is the measurement of discriminator which only uses the global features, which is not consistent with the hierarchical attention model in the generator side. As can be seen from the Table \ref{Tab:gan}, the results from this model are similar to that of global attention model, since the reward signal from the discriminator tends to force the generator to produce sentences that only matches the global features.

We further build a model exactly like in the one defined in Fig. \ref{gan}. This model includes both of the global image features and the local object features, and thus guarantees that the discriminator and the generator are utilizing the same information source. The final results can be seen in Table \ref{Tab:gan}, which outperform all of other experimental settings.

\begin{table*}[!t]
\caption{Comparison of image captioning results on the MSCOCO dataset with previous methods, where $^{1}$ indicates external information are used during the training process and $^{2}$ means that reinforcement learning is applied to optimize the model. }
 \centering
\begin{tabular}{|c|c|c|c|c|c|c|c|}
\hline
  \hline
  Methods  & BLEU-1 & BLEU-2 & BLEU-3 & BLEU-4 & METEOR & CIDEr & ROUGE-L \\
  \hline
  Google NIC \cite{vinyals2015show} & 66.6 & 46.1 & 32.9 & 24.6 & - & -  & - \\
  m-RNN \cite{mao2014deep} & 67 & 49 & 35 & 25 & - & - & - \\
  BRNN \cite{karpathy2015deep} & 64.2 & 45.1 & 30.4 & 20.3 & - & - & - \\
  MSR/CMU \cite{7298856} & - & - & - & 19.0 & 20.4 & - & - \\
  Spatial Attention \cite{xu2015show} & 71.8 & 50.4 & 35.7 & 25.0 & 23.0 & - & - \\
  gLSTM \cite{jia2015guiding} & 67.0 & 49.1 & 35.8 & 26.4 & 22.7 & 81.3 & -  \\
  GLA \cite{8031355} & 56.8 & 37.2 & 23.2 & 14.6 & 16.6 & 36.2 & 41.9 \\
  MIXER \cite{ranzato2015sequence} & - & - & - & 29.0 & - & - & - \\
  SCA-CNN-ResNet \cite{chen2017sca} & 71.9 & \textbf{54.8} & \textbf{41.1} & \textbf{31.1} & 25.0 & - & - \\
  \hline
  Semantic Attention$^{1}$ \cite{you2016image} & 70.9 & 53.7 & 40.2 & 30.4 & 24.3 & - & - \\
  DCC$^{1}$ \cite{7780377} & 64.4 & - & - & - & 21.0 & -  & - \\
  \hline
  RL with G-GAN$^{2}$ \cite{dai2017towards}  & - & - & 30.5 & 29.7 & 22.4 & 79.5 & 47.5 \\
  RL with Embedding Reward$^{2}$  \cite{ren2017deep} &  71.3 & 53.9 & 40.3 & 30.4 & 25.1 & \textbf{93.7} & 52.5 \\
  \hline
  Ours$^{2}$ & \textbf{73.036} & 53.688 & 39.069 & 28.551 & \textbf{25.324} & 92.449 & \textbf{56.539} \\
  \hline
  \hline
\end{tabular}
\label{Tab:cmp}
\end{table*}%

To prove the effectiveness of the proposed method, we compare our final results on the ``Karpathy'' test split with previously published results, which is shown in Table \ref{Tab:cmp}. We list most of the published results on the ``Karpathy'' split, which are grouped into three categories. The first category corresponds to various methods without external information and reinforcement learning. The best of them (SCA-CNN-ResNet) is the spatial and channel-wise attention model \cite{chen2017sca} in which both the spatial and channel-wise attention mechanisms are utilized for image captioning. The methods in the second group use extra information during the training of the model. For instance, Semantic Attention \cite{you2016image} utilizes rich extra data from social media to train the visual attribute predictor. Deep Compositional Captioning (DCC) \cite{7780377} generates extra data to prove its unique transfer capability. The third group corresponds to the reinforcement learning technique. RL with G-GAN \cite{dai2017towards} applies conditional GAN and policy gradient to generate image descriptions. Although their results on the evaluation metrics are not improved, they prove that the generated captions are more diverse and naturalistic. Embedding Reward \cite{ren2017deep} applies a policy network to generate captions and a value network to evaluate the reward. Additionally, they also apply advanced inference method called lookahead inference and beam search during testing. They achieve the current state-of-the-art results on the ``Karpathy'' split. Although we do not use any external knowledge and any advanced inference technique (including beam search, we use greedy search in all of our experiments), we achieve similar results to the current state-of-the-art methods (Embedding Reward \cite{ren2017deep} and SCA-CNN-ResNet \cite{chen2017sca}), with state-of-the-art results on three important metrics: BLEU-1, METEOR and ROUGE-L and lead other methods significantly.

\subsubsection{Qualitative Evaluation}

In addition to the quantitative evaluation using the standard metrics, we qualitatively evaluate the proposed model by visualization. Firstly, we plot some global attention maps corresponding to each generated words as shown in Fig. \ref{vis_global}. It is obvious in the figure that the attentive regions normally correspond with the semantic meaning of the generated word in each time step. Then we choose some examples to visualize the local attention weights on the detected objects, which are shown in Fig. \ref{vis_local}. We only retrieve the top 10 detected objects and corresponding attentive weights obtained from the local attention mechanism because of limited space in the figure. The detector can detect some fine-grained objects, which provide complementary information for the global attention mechanism. At last, we show some of the generated sentences using different methods. Specifically, we show the ground-truth sentences, descriptions generated by the MLE training-based model and by the proposed model as shown in Fig. \ref{vis_caption}. The text in red are the sentences generated by the proposed model, which are more accurate and naturalist than the MLE-based model, which are shown in blue. Specially, the proposed model show superior performance in finding the fine-grained properties of the image since the RL model automatically measure the coherence of the sentences and the image content. For instance, in Fig. \ref{vis_caption} (c), the proposed model successfully determines the gender of the person in the image whilst the MLE training-based model gets it wrong.

\section{Conclusion}
This paper targets the image captioning task, which is a fundamental problem in artificial intelligence. Based on the recent successes of deep learning, especially the CNN feature representation and the LSTM with attention model, the paper proposes the use of a hierarchical attention mechanism, considering not only the global image features but also detected object features, with improved results. A significant improvement over the current RNN-based MLE training has also been demonstrated. Specifically, a GAN framework with RL optimization for the image captioning task is proposed to generate more accurate and high-quality captions. The discriminator is to evaluate the coherence and consistency between the generated sentences and image content, thus providing the rewards for optimization. The whole model follows a three-step training strategy. Experiments analysis confirms the merits of the framework and key contributors the improved performance. Comparable results with current state-of-the-art methods are achieved using only greedy inference, which proves the effectiveness of the training procedure.




%

\small
\bibliographystyle{IEEEtran}
\bibliography{action}

%




\end{document}